\documentclass[letterpaper]{article}
\usepackage{aaai2027}
\nocopyright
\usepackage[hyphens]{url}
\usepackage{graphicx}
\usepackage{natbib}
\usepackage{caption}
\usepackage{booktabs}
\usepackage{amssymb}
\newcommand{\benchname}{MMArch}
\newcommand{\yes}{\checkmark}
\newcommand{\no}{\textendash}
\title{MMArch: Benchmarking Multimodal Reasoning Grounded in Architectural Evidence}
\author{
    Chenxu Du\textsuperscript{\rm 2}\equalcontrib,
    Kang An\textsuperscript{\rm 1}\equalcontrib\thanks{Project leader},
    Tengyue Wang\textsuperscript{\rm 5},
    Zhongyu Yang\textsuperscript{\rm 3},
    Xinqi Yang\textsuperscript{\rm 6},\\
    Yuanchi Zhu\textsuperscript{\rm 7,\rm 8},
    Hebao Zhu\textsuperscript{\rm 9},
    Ziliang Wang\textsuperscript{\rm 4},
    Faqiang Qian\textsuperscript{\rm 4},
    Yunli Yang\textsuperscript{\rm 10},
    Qibing Ren\textsuperscript{\rm 1}\corresponding
}
\affiliations{
    \textsuperscript{\rm 1}Shanghai Jiao Tong University,
    \textsuperscript{\rm 2}Southwest Jiaotong University,
    \textsuperscript{\rm 3}ModelBest,
    \textsuperscript{\rm 4}SenseTime,\\
    \textsuperscript{\rm 5}South China University of Technology,
    \textsuperscript{\rm 6}East China Normal University,
    \textsuperscript{\rm 7}ShanghaiTech University,\\
    \textsuperscript{\rm 8}Institute of Automation, Chinese Academy of Sciences,
    \textsuperscript{\rm 9}Chongqing University,\\
    \textsuperscript{\rm 10}Institute for Advanced Algorithms Research, Shanghai\\
    dcx\_swjtu@outlook.com,
    \{an\_kang, renqibing\}@sjtu.edu.cn
}

\begin{document}
\maketitle

\begin{abstract}
Multimodal large language models (MLLMs) perform strongly on engineering imagery, yet existing benchmarks mostly test drawing recognition, information extraction, or compliance checking, leaving open whether models can combine distributed visual evidence with engineering principles to reach a conclusion. We introduce \benchname{}, a benchmark for architecture and civil engineering spanning ten subdomains and built entirely from figures in peer-reviewed papers. Its $1{,}212$ short-answer items are produced by a decoupled planner--writer pipeline and validated through automated screening, a blind adversarial audit, and expert review, so that answering requires perceiving the relevant evidence, identifying the governing principle, and applying it, not exploiting textual or single-figure shortcuts. Evaluating $18$ open-weight and proprietary MLLMs against a domain-expert panel, we find a wide gap: the strongest open-source model attains about $30\%$ and the best proprietary system $52\%$, while human experts reach $95\%$, more than forty points ahead. Our error analysis shows that failures concentrate in applying principles and combining evidence across figures rather than in locating it, pointing to substantial headroom for future research. Code and data are available at \url{https://dcx-swjtu.github.io/MMArch/}.
\end{abstract}

\begin{figure*}[t]
\centering
\includegraphics[width=\textwidth]{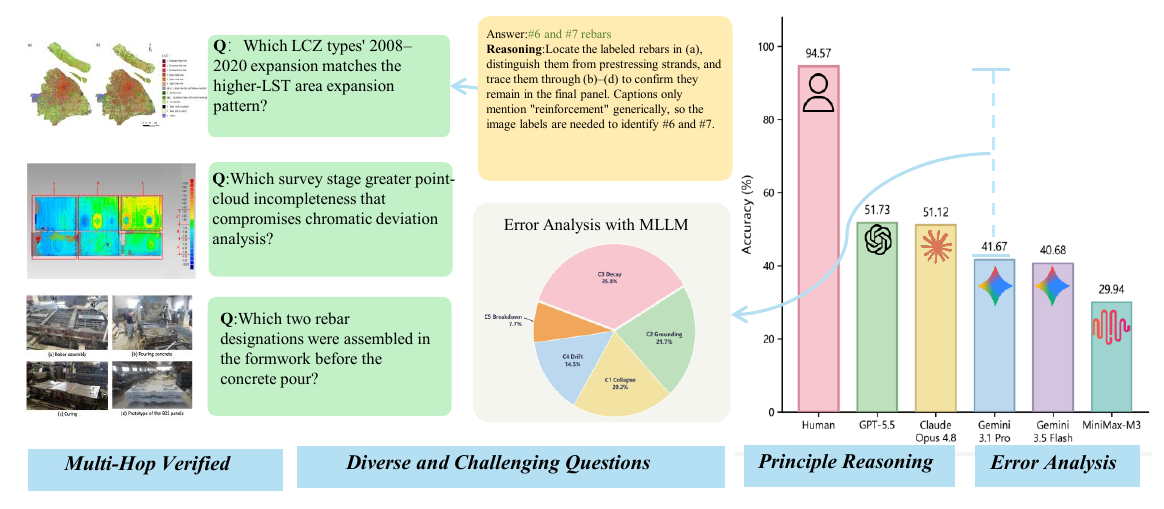}
\caption{Overview of \benchname{}. Left: representative items that require
  reading evidence jointly across figure panels and applying the governing
  engineering principle to answer; for one item we also show the underlying
  reasoning, included only to illustrate why the item is diagnostic and never
  exposed to evaluated models. Center: aggregate distribution of error categories
  from our error analysis. Right: accuracy (\%) of a human panel against
  representative MLLMs on \benchname{}; every evaluated model trails humans
  by more than forty points.}
\label{fig:teaser}
\end{figure*}

\section{Introduction}
Architecture, engineering, and construction (AEC) is one of the world's largest
economic sectors, and within it value is created less by physical building than
by the design--analysis--review cycle---among the most expertise-bound and
costly stages of the entire R\&D pipeline, so that even partial automation
carries outsized value. Professionals in this cycle spend much of their
effort interpreting technical figures (drawings, load--displacement curves,
thermal sections, BIM views), cross-referencing them, and applying codified
principles to judge whether a design is safe and compliant. This figure-grounded
reasoning recurs across the entire end-to-end AEC pipeline: from \emph{urban
planning} and \emph{spatial analysis}, through \emph{structural},
\emph{seismic}, \emph{building-physics}, \emph{envelope}, and
\emph{computational} design, to \emph{BIM}-based digital delivery and downstream
\emph{building inspection} and \emph{heritage conservation}, making it a natural
target for multimodal AI.

Multimodal large language models (MLLMs) have rapidly advanced in reading
and reasoning over technical imagery, positioning them to support AEC practice.
A core capability
this role requires is principle-grounded visual reasoning: combining evidence
across figures under a governing engineering principle, rather than reading any
single view alone. To steer progress toward this
capability, the community needs rigorous, diverse, and challenging evaluation
benchmarks.

However, most existing AEC benchmarks probe only single-figure perception or textual
knowledge recall, such as parsing floor plans and CAD drawings, or testing code knowledge,
document rules, and agentic workflows in text (Section~\ref{sec:related_work}). None of
them requires combining visual evidence with an explicit engineering principle to reach
a conclusion, as real-world judgment demands. Without a benchmark that isolates this
joint requirement, we cannot reliably measure or improve an MLLM's professional
reasoning competence.

In this work, we introduce \benchname{}, a benchmark dedicated to principle-grounded
multimodal reasoning in architecture and civil engineering
(Figure~\ref{fig:teaser}). \benchname{} decomposes this reasoning into three functional
stages, \textsc{Perceive}, \textsc{Know}, and \textsc{Judge}, that extract visual
evidence, identify the governing principle, and apply it to derive the answer; every
item requires all three, which serve as diagnostic lenses for error analysis rather
than as task labels. \benchname{} spans ten subdomains, from structural and seismic
engineering to urban planning and computational design (Figure~\ref{fig:subdomains}).

Crafting such a benchmark is far from trivial: prompting a single model to produce both
question and answer from a figure, as prior work does, leaves the ground truth prone to
hallucination. We instead adopt a fully decoupled pipeline: a planner selects a passage
and freezes an answer of at most ten tokens before a writer composes a free-form
question around it, so neither role can tailor its output to the other. Automated
screening then discards candidates solvable from text or captions alone or too easily
from the full input; three blind agents audit each survivor for evidence visibility,
principle necessity, and shortcuts; and three professional architects and engineers
retain an item only by unanimous agreement.

Screening roughly $10{,}000$ peer-reviewed papers from SCI-indexed journals and arXiv,
we distill $1{,}212$ validated items, each anchored to a specific principle and its
supporting evidence. We evaluate $18$ open-weight and proprietary MLLMs against a panel
of domain experts: the strongest open-source model attains about $30\%$, the
best proprietary system (GPT-5.5) only $51.7\%$, and the human panel $94.6\%$, a
gap of more than forty points.

This gap also reveals where progress is needed. Our error analysis groups failures into
five categories, perception, principle, composition, grounding, and consistency errors,
with composition error alone accounting for more than a third of failures, indicating
that combining evidence and principle, rather than locating it, is the primary
bottleneck. Simple prompting interventions such as chain-of-thought reasoning yield only
small, inconsistent gains, pointing to gaps in domain knowledge and compositional
reasoning rather than an absence of explicit reasoning traces.

\section{Related Work}
\label{sec:related_work}

\begin{table*}[t]
\centering
\footnotesize
\setlength{\tabcolsep}{4.5pt}
\renewcommand{\arraystretch}{1.12}
\begin{tabular*}{\textwidth}{@{\extracolsep{\fill}}l r c c c c c c c c c@{}}
\toprule
\textbf{Benchmark} & \textbf{\#Items} & \textbf{\#Dom.} & \textbf{Paper} & \textbf{Multi-} & \textbf{Open} & \textbf{Prin-} & \textbf{Anti-} & \textbf{Blind} & \textbf{Dual} & \textbf{Rule} \\
                   &                  &                 & \textbf{Fig.}  & \textbf{Fig.}   & \textbf{Ans.} & \textbf{ciple} & \textbf{cut}    & \textbf{Audit} & \textbf{Verif.} & \textbf{Score} \\
\midrule
\multicolumn{11}{@{}l}{\textit{Drawing \& CAD perception}} \\
CubiCasa5K~\cite{kalervo2019cubicasa}          & 5{,}000   & 1  & \no  & \no  & \no  & \no  & \no  & \no  & \no  & \yes \\
FloorPlanCAD~\cite{fan2021floorplancad}        & 15{,}000  & 1  & \no  & \no  & \no  & \no  & \no  & \no  & \no  & \yes \\
ArchCAD-400K~\cite{luo2025archcad}             & 400{,}000 & 1  & \no  & \no  & \no  & \no  & \no  & \no  & \no  & \yes \\
AECV-Bench~\cite{kondratenko2026aecvbench}     & 312       & 2      & \no & \no & \no  & \no  & \no  & \no  & \no  & \yes \\
\midrule
\multicolumn{11}{@{}l}{\textit{Professional knowledge \& documents}} \\
AECBench~\cite{liang2026aecbench}              & 4{,}800 & 23     & \no  & \no  & \yes & \no  & \no  & \no  & \no  & \no  \\
DesignQA~\cite{doris2025designqa}             & 1{,}451 & 6      & \no  & \no  & \yes & \yes & \no  & \no  & \no  & \no  \\
TechMB~\cite{kunz2025techmb}                  & 947     & 1      & \no  & \no  & \yes & \no  & \no  & \no  & \no  & \yes \\
\midrule
\multicolumn{11}{@{}l}{\textit{Agentic workflows}} \\
AEC-Bench~\cite{mankodiya2026aecbench}        & 196     & 9      & \no  & \no  & \yes & \no  & \no  & \no  & \no  & \no  \\
\midrule
\multicolumn{11}{@{}l}{\textit{Scientific-figure QA}} \\
CharXiv~\cite{wang2024charxiv}                & 2{,}323 & 8  & \yes & \no  & \yes & \no  & \no  & \no  & \no  & \no  \\
MMMU-Pro~\cite{yue2024mmmupro}               & 3{,}460 & 30 & \no  & \no  & \no  & \yes & \yes & \no  & \no  & \yes \\
\midrule
\textbf{\benchname{} (ours)}                 & 1{,}212 & 10 & \yes & \yes & \yes & \yes & \yes & \yes & \yes & \yes \\
\bottomrule
\end{tabular*}
\caption{\benchname{} versus representative AEC and scientific-figure
  benchmarks. \emph{Paper Fig.}: items built from real research-paper figures;
  \emph{Multi-Fig.}: reasoning jointly across multiple figures; \emph{Open
  Ans.}: open-ended (non-multiple-choice) answers; \emph{Principle}: answering
  requires applying a domain principle to the visual evidence;
  \emph{Anti-cut}: text-/caption-only shortcut screening; \emph{Blind Audit}:
  blind adversarial audit; \emph{Dual Verif.}: independent two-path answer
  verification; \emph{Rule Score}: deterministic rule-based scoring
  (vs.\ an LLM judge). \yes~= yes, \no~= no. \benchname{} is the only benchmark
  that couples cross-figure, principle-grounded reasoning with a full
  construction-time quality stack.}
\label{tab:bench_compare}
\end{table*}

\paragraph{Benchmarks for the AEC domain.} Existing AEC benchmarks evaluate several distinct capabilities, spanning perceptual parsing, professional-knowledge recall, and agentic workflows (Table~\ref{tab:bench_compare}). One line of work centers on perceptual parsing of drawings: CubiCasa5K, FloorPlanCAD, and ArchCAD-400K parse, count, and recognize symbols in floor plans and CAD drawings \cite{kalervo2019cubicasa,fan2021floorplancad,luo2025archcad}, and AECV-Bench extends similar perceptual evaluation to general-purpose multimodal models \cite{kondratenko2026aecvbench}. A second line probes professional knowledge and document use rather than figure interpretation: AECBench examines building-code knowledge in text \cite{liang2026aecbench}, DesignQA evaluates rule extraction and application from engineering documents \cite{doris2025designqa}, and TechMB tests the interpretation of individual mechanical drawings \cite{kunz2025techmb}. A third, AEC-Bench, evaluates agentic systems in project-level workflows \cite{mankodiya2026aecbench}. These efforts are complementary to our work, but none explicitly isolates whether a model can select relevant evidence from professional figures, connect that evidence through architectural principles, and derive a problem-specific conclusion. \benchname{} closes this gap with open-ended questions, grounded in figures from the AEC literature, that require exactly this combination of evidence selection and principled synthesis.

\paragraph{QA over scientific figures.} A parallel line of work studies question answering over scientific graphics, but largely treats relevance as given rather than as something a model must discover. Early benchmarks were built from synthetic plots \cite{kahou2017figureqa,methani2020plotqa}, followed by real-world charts \cite{masry2022chartqa}, documents and diagrams \cite{mathew2021docvqa,mathew2022infographicvqa,kembhavi2016ai2d}, figures from scientific papers \cite{wang2024charxiv,pramanick2024spiqa,roberts2024scifibench,li2024mmsci}, and exam-style multimodal problems \cite{lu2022scienceqa,lu2024mathvista,yue2024mmmu}. These benchmarks have substantially advanced visual and scientific reasoning, yet most questions still carry explicit or implicit cues about which visual elements, panels, or operations matter. In architectural reasoning, identifying the governing evidence is itself part of the challenge: the model must determine which condition controls, which response is critical, and which principle connects the observations to the conclusion. \benchname{} therefore treats evidence selection as part of the reasoning problem rather than as information supplied by the prompt.

\paragraph{Shortcut-resistant benchmark construction.} A separate line of work targets construction methodology itself, seeking to keep benchmarks free of the shortcuts that let models bypass genuine reasoning. MME, MMBench, and SEED-Bench rely on careful manual curation \cite{fu2023mme,liu2024mmbench,li2023seedbench}; MMStar removes samples answerable by vision-blind models \cite{chen2024mmstar}; and MMMU-Pro strengthens its predecessor by filtering text-solvable questions, expanding answer options, and embedding questions into images \cite{yue2024mmmupro}. Adversarial and model-assisted data construction has also been explored in SWAG, HellaSwag, ANLI, and Dynabench \cite{zellers2018swag,zellers2019hellaswag,nie2020anli,kiela2021dynabench}. \benchname{} builds on these practices by decoupling answer generation from question construction through a planner--writer pipeline with answer freezing, and by combining text-only and caption-only screening with a blind three-agent adversarial audit to strip out linguistic shortcuts and confirm that both visual evidence and domain knowledge are required. Two further safeguards go beyond this prior practice: every surviving answer is independently re-derived along two separate paths and reconciled before acceptance, and every item must earn unanimous agreement from three professional architects and engineers under a constrained, deterministically scored short-answer format that admits neither multiple-choice guessing nor an LLM judge.

\section{The \benchname{} Benchmark}

In this section, we present \benchname{}, a benchmark for evaluating the ability
of MLLMs to draw domain-specific conclusions by jointly interpreting
visual evidence and applying architectural or engineering
knowledge. We first define the task in Section~\ref{sec:mmarch_overall}, then
describe the construction pipeline in Section~\ref{sec:mmarch_construction}.

\subsection{Overview of \benchname{}}
\label{sec:mmarch_overall}

This principle-grounded reasoning capability involves extracting case-specific
observations from one or more AEC figures, identifying the applicable domain
knowledge, and synthesizing the two to derive a problem-specific conclusion.
To evaluate it in multimodal large language models, we introduce \benchname{}.

Formally, each item is represented as \(x=(\mathcal{I},Q,A)\), where
\(\mathcal{I}\) contains one to three figures from a single peer-reviewed
paper, \(Q\) is a short-answer question, and \(A\) is the reference answer.
The source-paper passages used to establish answer provenance are retained as
curator-only metadata and are not exposed to models being evaluated.

We decompose this reasoning process into three functional stages:
\textsc{Perceive}, \textsc{Know}, and \textsc{Judge}.
\textsc{Perceive} extracts, compares, and associates relevant evidence within
or across figures;
\textsc{Know} identifies the engineering principle, convention, constraint, or
workflow rule applicable to the observed case; and
\textsc{Judge} applies this knowledge to the extracted visual observations to
derive the answer.
These stages are not disjoint task categories: every valid item requires all
three, and the answer cannot be determined if either the required visual
evidence or the applicable AEC knowledge is absent. Knowledge necessity, in
particular, is verified through independent review and expert audit during
construction. The stages serve as
lenses for task design and error diagnosis.

We screened roughly $10{,}000$ papers from SCI-indexed journals and from
arXiv, extracted candidate figures, and passed them through a multi-stage
construction pipeline that authors and validates each item while enforcing
correctness, figure necessity, and reasoning-based difficulty
(Section~\ref{sec:mmarch_construction}). This yields $1{,}212$ validated short-answer
items, each anchored to a specific engineering principle and its supporting
visual evidence.

\begin{figure}[t]
\centering
\includegraphics[width=0.85\columnwidth]{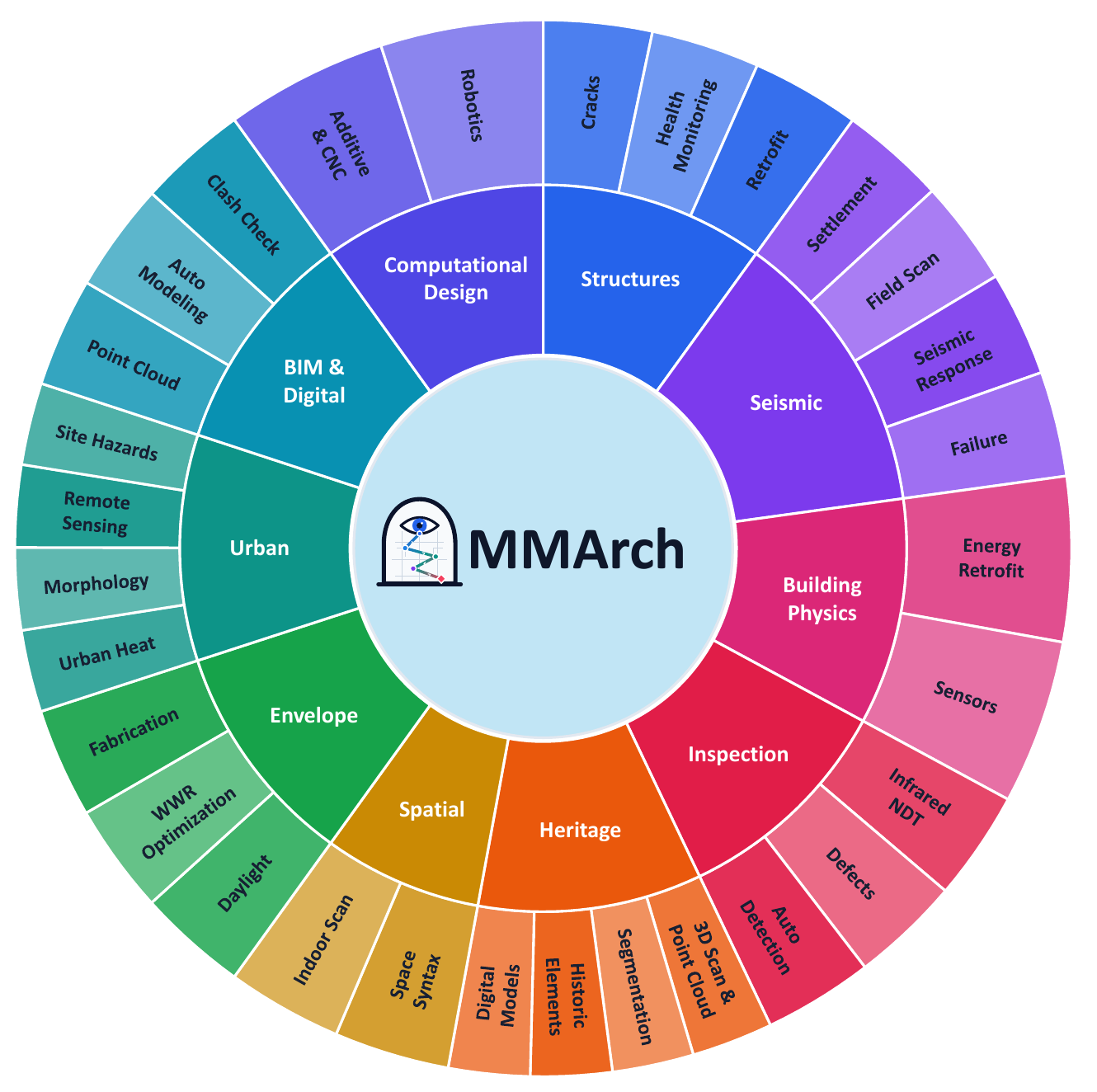}
\caption{The ten subdomains covered by \benchname{}, with per-subdomain item
  counts and proportions.}
\label{fig:subdomains}
\end{figure}

The benchmark covers ten subdomains spanning architecture and civil
engineering~(Figure~\ref{fig:subdomains}):
structural engineering~(Struct.), seismic engineering~(Seism.), building
physics~(Phys.), building inspection~(Insp.), heritage
conservation~(Herit.), spatial analysis~(Spat.), building-envelope
engineering~(Envlp.), urban planning~(Urban), BIM and digital
construction~(BIM), and computational design~(CDes.).
Figure~\ref{fig:subdomains} also reports the item count and relative
proportion for each subdomain.

\subsection{Benchmark Construction Process}\label{sec:mmarch_construction}

\begin{figure*}[t]
\centering
\includegraphics[width=\textwidth]{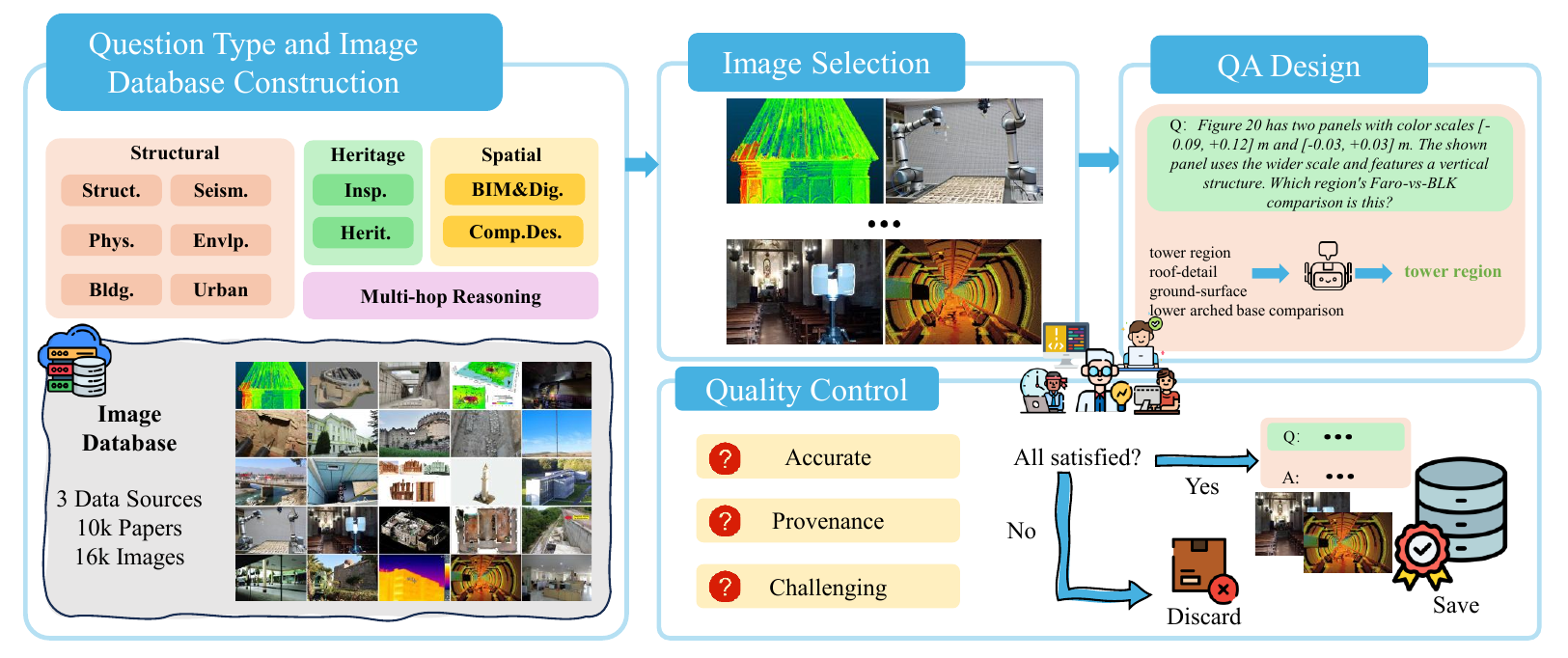}
\caption{Illustration of the \benchname{} construction pipeline: images
  are collected from peer-reviewed academic papers, relevant image sets
  are carefully selected, complex QA tasks are constructed through a
  decoupled planner--writer design, and all data undergo rigorous
  quality control.}
\label{fig:pipeline}
\end{figure*}

A central design goal of \benchname{} is that item quality is enforced during
construction rather than checked after the fact. The pipeline is organized
around the three properties that make an item diagnostic: its answer must be
correct and grounded in the source paper, its figures must be necessary, and
its difficulty must arise from reasoning rather than
illegibility or ambiguity. The pipeline realizes these properties through a
sequence of stages (Figure~\ref{fig:pipeline}), each targeting a specific
failure mode; a candidate is only ever passed or discarded, never patched, so
that no item is retained on the strength of a corrected defect. We treat the resulting low yield as the
price of this rigor rather than a quantity to be optimized.

\paragraph{Data Collection.}
Evaluating professional reasoning requires visual evidence as dense and
authentic as the material that AEC researchers and engineers interpret in
practice; this
precludes synthetic or illustrative imagery. We therefore build \benchname{}
entirely from figures in the peer-reviewed AEC literature, where a single figure
often conveys the quantitative basis of an engineering finding. We collect
roughly $10{,}000$ papers from SCI-indexed journals and arXiv across the
ten subdomains shown in Figure~\ref{fig:subdomains}, and keep only figures that
present measurable or diagnostic content, excluding decorative and purely
illustrative graphics. For each figure, we retain its full context, comprising
the image, the caption, and the surrounding text. The source passages that
support an answer are stored as curator-only metadata and are not made available
to the models being evaluated. Finally, for every figure we identify the sentences in
the paper that explicitly discuss it, which we refer to as \emph{evidence
leads}. These leads anchor each answer to the published text, and we give
preference to comparative and relational statements over descriptive ones, so
that answering a question requires interpreting the figure rather than restating
its caption.

\paragraph{Question-Answer Design.}
A common practice in benchmark construction is to prompt a single model to
produce both the question and its answer from a
figure~\cite{li2023seedbench,liu2024mmbench}. Even when the resulting pairs are
filtered by human annotators, this coupling remains problematic: because the
model decides what to ask and what the answer is at once, the ground truth is
prone to hallucination and internal inconsistency, and manual verification is
difficult to scale.

To remove this failure mode, we decouple answer generation from question
construction across two specialized agents. A \emph{planner} first selects an
evidence lead and distills the answer from that lead into a span of at most ten
tokens, together with the visual observations and intermediate steps that
support it; the answer is then frozen and cannot be altered downstream. A
\emph{writer} then composes a free-form question around this frozen answer;
templates are avoided so that phrasing stays varied and cannot be
pattern-matched. Each question must satisfy three requirements: the answer must
never appear in it; reaching it must require at least two visual operations on
the figure rather than a single lookup; and the question must be accurate,
unambiguous, and answerable in principle by a domain expert. The writer is
further instructed to make each question as challenging as possible for current
MLLMs while keeping it well-posed. Because the answer is fixed before the
question exists and the two roles remain separate, neither can be silently
tailored to the other. Each surviving candidate then enters the
automated validation and independent review described below.

\paragraph{Quality Control.}
Structural validity does not ensure that an item is difficult, unambiguous, or
correctly answered. Each candidate therefore passes three automated checks and a
final expert review before release.

With reasoning disabled, we query a solver on the full input ($8$ trials), the
question alone ($4$), and the caption only ($2$); two reasoning-enabled trials
are kept for diagnosis. An item solved in any text- or caption-only trial
(leakage), or in at least five of the eight full-input trials (too easy), is
discarded; the rest are labeled \emph{Hard} ($2$--$4/8$) or \emph{Challenge}
($1/8$), while $0/8$ items are deferred to expert review.

Three agents, each blind to the answer and the scores, then audit the item: one
checks that the required evidence is visible and that the cited source span
matches; one confirms that a solution needs all three stages (perception,
knowledge, and judgment) rather than a shortcut; one searches adversarially for
shortcuts, alternative answers, and ambiguity. Any fatal finding rejects the
item; uncertain cases are deferred to expert review.

We also cross-check the answer along two paths, one recovered from the cited
paper span and one re-derived from image crops by a separate curator-side
computation, keeping the item only if both agree under a fixed unit, rounding,
and tolerance. Finally, three professional architects and engineers review all
remaining and deferred items and retain an item only by unanimous agreement;
after deduplication across papers, figures, and principles, $1{,}212$ items
remain.

\section{Evaluation on \benchname{}}

\subsection{Evaluation Setup}

We evaluate a broad range of MLLMs on \benchname{}
(Table~\ref{tab:main_results}): open-source models spanning the Qwen, Gemma,
InternVL, and Llama families together with MiniMax M3, and
proprietary models including GPT-4o, GPT-5.5, Claude Opus~4.8, Claude Sonnet 4.5,
Gemini~3.1 Pro Preview, and Gemini~3.5 Flash. For consistency, every
model is run with a decoding temperature of $0$ and a maximum output length of
$2{,}048$ tokens. Proprietary models were queried through their official APIs
on 1~July 2026, while open-weight models were served from their publicly
released checkpoints with vLLM through the latest release of LLaMA-Factory on
NVIDIA~A100 GPUs; because decoding is deterministic, each reported number
reflects a single run. As an upper reference, we additionally report a \emph{Human}
baseline, measured as the average accuracy of a panel of qualified architects
and engineers who did not participate in benchmark construction.
Because \benchname{} poses short-answer rather than multiple-choice questions,
we report accuracy (\%) as the match between the final answer extracted from a
model output and the reference answer, under a normalization that accounts for
units, synonyms, numeric tolerance, and accepted answer sets.

\subsection{Main Results}
\begin{table*}[t]
\centering
\footnotesize
\setlength{\tabcolsep}{3pt}
\begin{tabular*}{\textwidth}{@{\extracolsep{\fill}}l rrrrrrrrrrr@{}}
\toprule
\textbf{Model}                        & \textbf{Struct.} & \textbf{Seism.} & \textbf{Phys.} & \textbf{Insp.} & \textbf{Herit.} & \textbf{Spat.} & \textbf{Envlp.} & \textbf{Urban} & \textbf{BIM} & \textbf{CDes.} & \textbf{Avg.} \\
\midrule
\multicolumn{12}{@{}l}{\textit{Open-source models}} \\
Qwen3.6-27B                           &  17.33 &  25.33 &  18.33 &  24.33 &  14.33 &  23.33 &  27.33 &  16.33 &  21.33 &  15.33 &  20.33 \\
Qwen3.6-35B-A3B                       &  18.78 &\textit{27.18}&  19.83 &  26.13 &  15.63 &\textit{25.08}&\textit{29.28}&  17.73 &  22.98 &  16.68 &  21.93 \\
Qwen3.5-9B                            &  17.48 &  11.04 &  18.40 &  23.00 &  12.88 &  10.12 &  16.56 &  11.96 &  22.08 &  12.88 &  15.64 \\
Qwen3.5-27B                           &  21.05 &  14.05 &  22.05 &  27.05 &  16.05 &  13.05 &  20.05 &  15.05 &  26.05 &  16.05 &  19.05 \\
Qwen3.5-35B-A3B                       &  21.90 &  14.62 &  22.94 &  28.14 &  16.70 &  13.58 &  20.86 &  15.66 &  27.10 &  16.70 &  19.82 \\
Gemma-4-31B                           &\textit{28.35}&  22.35 &  30.35 &\textit{30.25}&  32.35 &  16.35 &  24.35 &  29.35 &  31.35 &  13.45 &  25.85 \\
Qwen2.5-VL-7B-Instruct                &  19.05 &   9.55 &  15.25 &   8.60 &  16.20 &  20.00 &   7.65 &  17.15 &  10.50 &   9.55 &  13.35 \\
InternVL3-8B                          &  20.71 &  10.41 &  16.59 &   9.38 &  17.62 &  21.74 &   8.35 &  18.65 &  11.44 &  10.41 &  14.53 \\
Llama-3.2-11B-Vision-Instruct         &  11.37 &  16.77 &  21.27 &  10.47 &  15.87 &   9.57 &  17.67 &  22.17 &  13.17 &  11.37 &  14.97 \\
Llama-3.2-90B-Vision-Instruct         &  15.08 &  21.38 &  26.63 &  14.03 &  20.33 &  12.98 &  22.43 &  27.68 &  17.18 &  15.08 &  19.28 \\
Llama-4-Scout-17B-16E-Instruct        &  16.74 &  23.34 &  28.84 &  15.64 &  22.24 &  14.54 &  24.44 &  29.94 &  18.94 &  16.74 &  21.14 \\
MiniMax M3                             &  27.72 &  24.90 &\textit{34.07}&  28.27 &\textit{34.21}&  25.00 &  27.66 &\textit{32.39}&\textit{31.89}&\textit{33.33}&\textit{29.94} \\
\midrule
\multicolumn{12}{@{}l}{\textit{Proprietary models}} \\
GPT-5.5                               &\textbf{58.42}&\textbf{52.17}&\textbf{48.93}&\textbf{55.61}&\textbf{44.28}&\textbf{61.35}&\textbf{50.74}&\textbf{53.89}&\textbf{49.56}&\textbf{42.18}&\textbf{51.73} \\
Claude Opus 4.8                       &  57.15 &  51.83 &  47.62 &  54.90 &  43.75 &  60.28 &  49.31 &  52.46 &  48.93 &  41.05 &  51.12 \\
Claude Sonnet 4.5                     &  52.38 &  47.65 &  44.19 &  50.72 &  40.36 &  55.84 &  45.67 &  48.93 &  45.28 &  38.42 &  46.94 \\
Gemini 3.1 Pro Preview                &  46.83 &  42.15 &  39.74 &  45.28 &  36.52 &  48.91 &  40.35 &  43.67 &  40.15 &  34.28 &  41.67 \\
Gemini 3.5 Flash                      &  45.62 &  41.38 &  38.95 &  44.17 &  35.84 &  47.53 &  39.62 &  42.85 &  39.47 &  33.56 &  40.68 \\
GPT-4o                                 &  38.45 &  35.72 &  33.18 &  37.64 &  29.85 &  41.26 &  32.94 &  35.83 &  31.47 &  27.56 &  34.27 \\
\midrule
\multicolumn{12}{@{}l}{\textit{Baseline}} \\
Human                                  &  89.25 &  95.92 &  91.85 &  93.21 &  97.37 &  91.67 &  91.49 & 100.00 &  97.30 &  96.67 &  94.57 \\
\bottomrule
\end{tabular*}
\caption{Accuracy (\%) across ten AEC subdomains. Per column, the best model
score is in \textbf{bold} and the best open-source score in \textit{italic};
Human is an upper-reference baseline.}
\label{tab:main_results}
\end{table*}

Table~\ref{tab:main_results} summarizes the results. The picture is consistent
across models: none exceeds $52\%$ average accuracy, and the strongest systems,
GPT-5.5 ($51.73\%$) and Claude Opus~4.8 ($51.12\%$), remain more than forty
points below the human panel at $94.57\%$. We highlight four observations.

\paragraph{The model--human gap reflects genuine reasoning, not shortcuts.} Because
each item is screened during construction to resist text-only and caption-only
solutions (Section~\ref{sec:mmarch_construction}), the large gap above cannot be
explained by language priors or caption reading; it isolates the joint visual
and domain reasoning that \benchname{} is designed to measure. The consistently
high human scores, between $89\%$ and $100\%$ across subdomains, further confirm
that the items are well-posed rather than noisy.

\paragraph{Proprietary models lead by a substantial margin over open-source
models.} Proprietary systems occupy the top of the table, while the strongest
open-source model, MiniMax M3 ($29.94\%$), trails the best proprietary model by
more than twenty points and falls short of every proprietary model evaluated
here. The remaining open-source models fall between roughly $13\%$ and $26\%$,
leaving substantial headroom for open models on professional visual reasoning.

\paragraph{Scaling model size brings limited gains.} Within a model family,
additional parameters help only marginally: the dense Llama-3.2 models improve
by about four points from 11B to 90B despite an eightfold increase in size, and
the dense Qwen3.5 models gain only about three points from 9B to 27B.
Performance on \benchname{} therefore appears bounded less by raw model
capacity than by whether a model can bind visual evidence to the governing
domain knowledge, a limitation that additional parameters, on their own, do
not resolve.

\paragraph{Difficulty varies across subdomains.} Subdomain-level difficulty is
uneven and depends on model class. Every proprietary model attains its highest
accuracy on spatial analysis, and its residual difficulty concentrates instead
in knowledge-intensive subdomains (computational design and heritage
conservation are the two weakest for all six proprietary models), where a
domain-specific principle, rather than perception, is the binding constraint.
Open-source models show no single universal weak spot: spatial analysis is the
weakest subdomain for the Qwen3.5 and Llama families and is among the weakest
for the strongest open model (MiniMax M3 scores just $25.00\%$ on spatial
analysis, against $31.89\%$ on BIM and digital construction), yet a few smaller
models peak on spatial instead. This divergence indicates that the source of
difficulty on \benchname{} shifts with model class rather than following a
single, universal ranking of subdomains. A plausible explanation is that
spatial-analysis items reward the general-purpose visual grounding that
large-scale proprietary pre-training develops well.
\subsection{Limited Effect of Prompting Techniques on \benchname{}}
\begin{figure}[t]
\centering
\includegraphics[width=\linewidth]{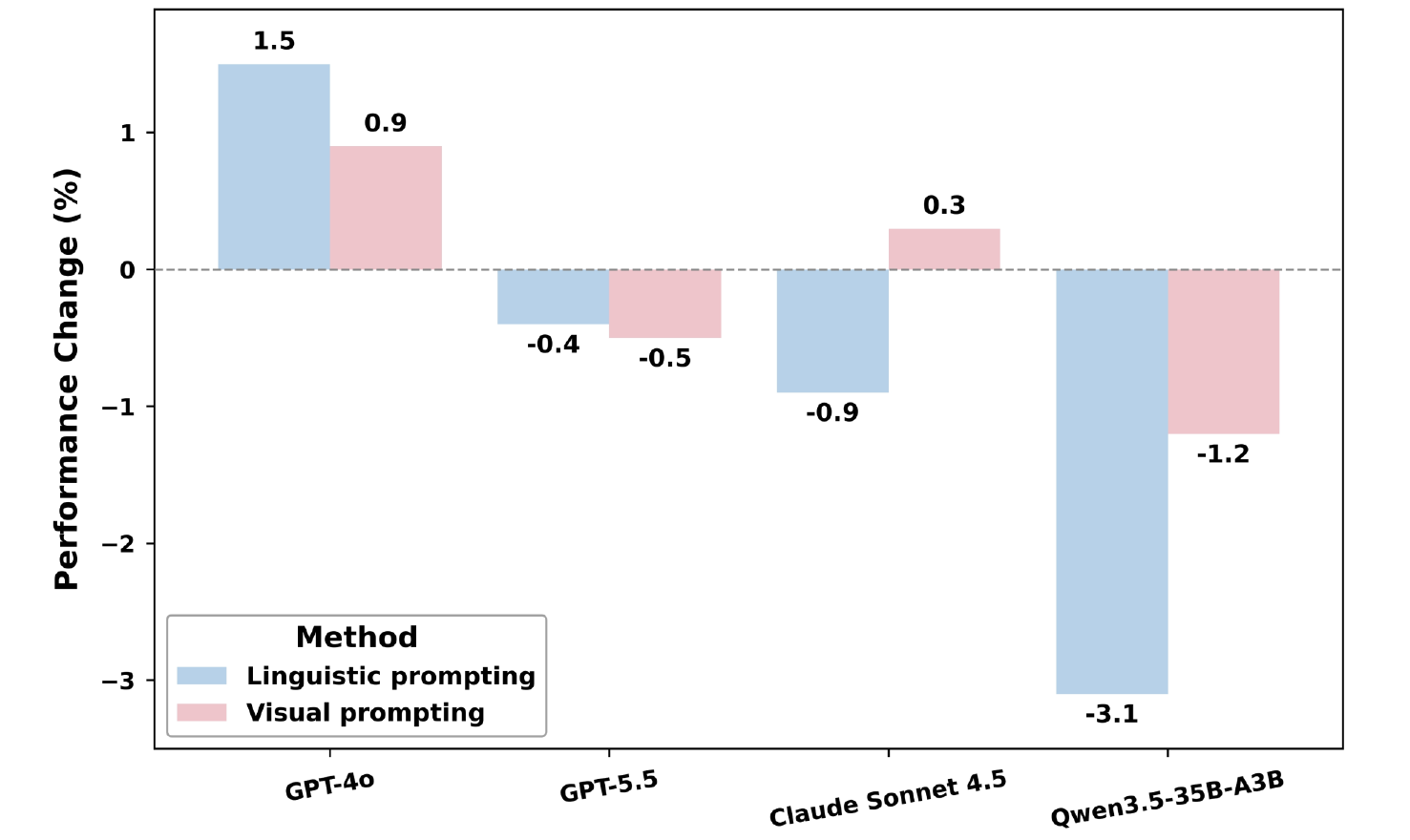}
\caption{Performance change (\%) from linguistic (chain-of-thought) and
visual (panel-identifier) prompting relative to the no-prompting baseline,
for four representative models.}
\label{fig:prompt}
\end{figure}

We test whether prompting techniques, widely used to improve reasoning in
large models, also help on \benchname{}, studying one linguistic and one
visual prompting strategy
on four representative models: GPT-4o, GPT-5.5, Claude Sonnet 4.5, and
Qwen3.5-35B-A3B. Figure~\ref{fig:prompt} reports the results.

\paragraph{Linguistic prompting.} We adopt Zero-Shot Chain-of-Thought (CoT),
prepending ``Let's think step by step'' to each prompt. CoT yields a
modest gain for GPT-4o ($+1.5$ points) but degrades performance for
the other three models, most severely for Qwen3.5-35B-A3B ($-3.1$
points). The limited and inconsistent effect suggests
that the bottleneck is not the absence of explicit reasoning traces;
rather, these models lack the domain-specific knowledge required to
construct valid chains.
\paragraph{Visual prompting.} We add panel identifiers (``[A]'', ``[B]'',
``[C]'') to each figure in concatenated multi-figure inputs, letting a
question refer to a specific figure directly, and compare against a baseline
without identifiers; because \benchname{} figures share no geometric overlap,
correspondence-based visual prompting is inapplicable here. Panel identifiers
produce small gains for GPT-4o and Claude Sonnet 4.5 ($+0.9$ and $+0.3$ points,
respectively) but degrade performance for GPT-5.5 and Qwen3.5-35B-A3B ($-0.5$
and $-1.2$ points).
The mixed pattern indicates that most models cannot exploit explicit
cross-figure binding cues: they either fail to associate the label
with the correct figure or, having made the association, still cannot
execute the multi-hop inference required to combine evidence across
figures. The difficulty therefore lies in compositional reasoning and
domain knowledge, not in visual alignment ambiguity.

\section{Error Analysis}
\label{sec:error_analysis}

\begin{figure}[t]
\centering
\includegraphics[width=\linewidth]{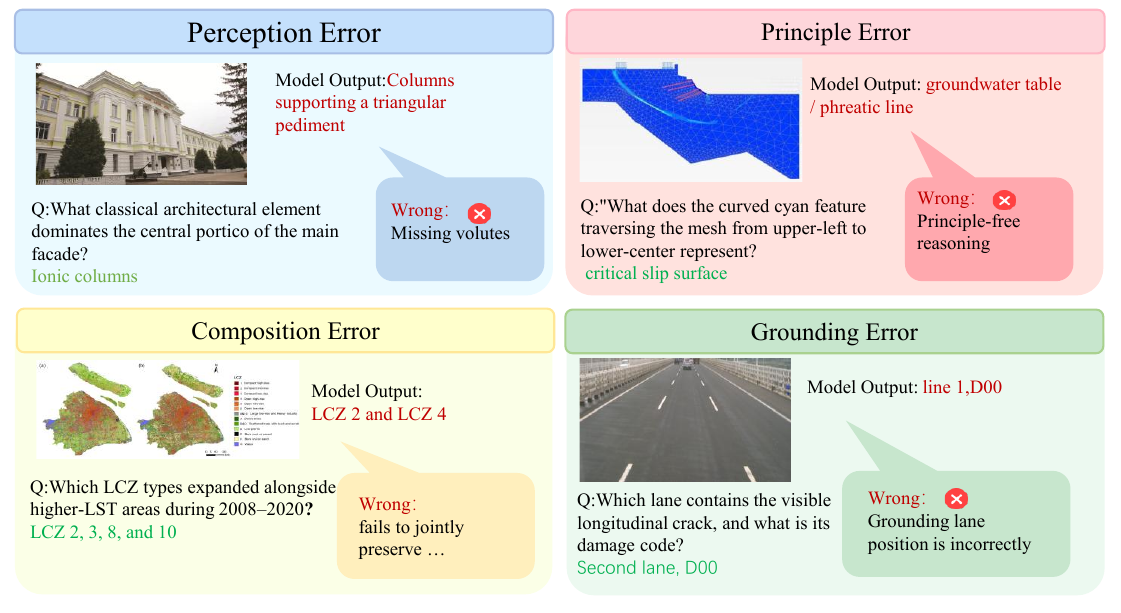}
\caption{Illustration of four error types identified in MLLM reasoning on
  \benchname{}.}
\label{fig:error_examples}
\end{figure}

\begin{figure}[t]
\centering
\includegraphics[width=\linewidth]{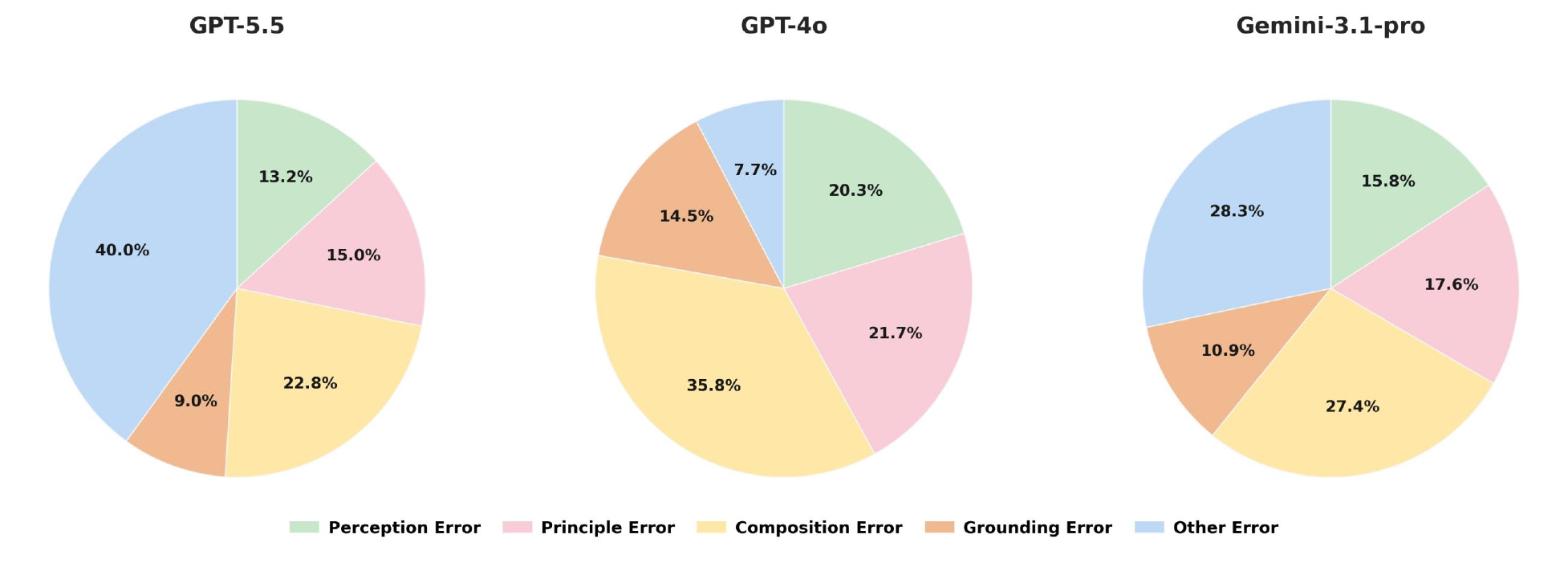}
\caption{Per-model distribution of correct answers and error types for three
  representative MLLMs, showing how the error mix shifts with model strength.}
\label{fig:error_by_model}
\end{figure}

To understand what limits performance beyond the aggregate accuracy in
Table~\ref{tab:main_results}, we manually inspect and categorize every model
failure on \benchname{} (the complete set rather than a sample), grouping
them into five categories, whose overall distribution is summarized in the
center panel of Figure~\ref{fig:teaser}. The percentages below give each
category's share of these failures, pooled across the evaluated models.
Figure~\ref{fig:error_examples} shows a representative failure case for four of
these categories, and Figure~\ref{fig:error_by_model} breaks the same
categories down per model, showing how the mix shifts with model strength.

\paragraph{Perception Error.} Making up $20.3\%$ of failures, these occur when
the visual signal is insufficient for a correct reading, e.g., mistaking the
most visually prominent curve among four overlapping load--displacement curves
for the correct one.

\paragraph{Principle Error.} In $21.7\%$ of failures, the evidence is read
correctly but an intuitive heuristic replaces the governing principle, e.g.,
inferring ``wider hysteresis loops mean softening'' where the correct
domain definition implies hardening.

\paragraph{Composition Error.} The dominant mode, at $35.8\%$ of failures:
every intermediate quantity is extracted correctly, but the final multi-step
combination drops or mishandles a term, e.g., omitting a base thermal
resistance and computing $60$~mm instead of the correct $80$~mm.

\paragraph{Grounding Error.} In $14.5\%$ of failures, the correct quantity is
identified but its value is misread or sign-flipped, e.g., reporting
$+138$~kN instead of $-142$~kN on an axial-force diagram.

\paragraph{Consistency Error.} The smallest category at $7.7\%$ of failures,
where the model's own analysis is correct but its stated answer is not, e.g.,
correctly identifying shear-failure indicators yet answering ``flexural
failure.''

Composition errors alone account for more than a third of failures, ahead of
principle and perception errors. Mapped onto the \textsc{Perceive},
\textsc{Know}, and \textsc{Judge} stages of Section~\ref{sec:mmarch_overall},
correctly combining evidence and principle is the largest bottleneck; the
remaining grounding and consistency errors (misreading a correctly identified
quantity, or contradicting one's own analysis) form a smaller tail.

The error mix also shifts with model strength (Figure~\ref{fig:error_by_model}):
as models improve on the dominant, well-characterized error modes, an
increasing share of their remaining failures falls outside these five
categories, suggesting that further progress will require diagnosing a
more heterogeneous long tail of mistakes rather than addressing a single
bottleneck.

\section{Conclusion}
We introduced \benchname{}, a benchmark of $1{,}212$ short-answer items built
entirely from figures in peer-reviewed architecture and civil-engineering
papers. A decoupled planner--writer pipeline with answer freezing, automated
shortcut screening, and unanimous expert review makes each answer require
perceiving evidence across figures, identifying the governing engineering
principle, and applying it, rather than exploiting linguistic or single-figure
shortcuts. Across $18$ open-weight and proprietary MLLMs, even the strongest
trails a domain-expert panel by more than forty points. This gap is driven less
by locating evidence than by compositional principle application, and
chain-of-thought and cross-figure prompting yield only small, inconsistent
gains. \benchname{} covers research-paper figures across ten AEC subdomains, and
thus does not probe field imagery or models beyond a current snapshot; even so,
the gap it reveals points to advances in model architecture or training rather
than prompting, and we intend it as a diagnostic testbed for that progress.

\bibliography{related}

\begin{thebibliography}{30}
\providecommand{\natexlab}[1]{#1}

\bibitem[{Chen et~al.(2024)Chen, Li, Dong, Zhang, Zang, Chen, Duan, Wang, Qiao,
  Lin, and Zhao}]{chen2024mmstar}
Chen, L.; Li, J.; Dong, X.; Zhang, P.; Zang, Y.; Chen, Z.; Duan, H.; Wang, J.;
  Qiao, Y.; Lin, D.; and Zhao, F. 2024.
\newblock Are We on the Right Way for Evaluating Large Vision-Language Models?
\newblock In \emph{Advances in Neural Information Processing Systems
  (NeurIPS)}, volume~37.

\bibitem[{Doris et~al.(2025)Doris, Grandi, Tomich, Alam, Ataei, Cheong, and
  Ahmed}]{doris2025designqa}
Doris, A.~C.; Grandi, D.; Tomich, R.; Alam, M.~F.; Ataei, M.; Cheong, H.; and
  Ahmed, F. 2025.
\newblock {DesignQA}: A Multimodal Benchmark for Evaluating Large Language
  Models' Understanding of Engineering Documentation.
\newblock \emph{Journal of Computing and Information Science in Engineering},
  25(2): 021009.

\bibitem[{Fan et~al.(2021)Fan, Zhu, Li, Chen, Zhu, and
  Tan}]{fan2021floorplancad}
Fan, Z.; Zhu, L.; Li, H.; Chen, X.; Zhu, S.; and Tan, P. 2021.
\newblock {FloorPlanCAD}: A Large-Scale {CAD} Drawing Dataset for Panoptic
  Symbol Spotting.
\newblock In \emph{Proceedings of the IEEE/CVF International Conference on
  Computer Vision (ICCV)}, 10128--10137.

\bibitem[{Fu et~al.(2025)Fu, Chen, Shen, Qin, Zhang, Lin, Yang, Zheng, Li, Sun,
  Wu, Ji, Shan, and He}]{fu2023mme}
Fu, C.; Chen, P.; Shen, Y.; Qin, Y.; Zhang, M.; Lin, X.; Yang, J.; Zheng, X.;
  Li, K.; Sun, X.; Wu, Y.; Ji, R.; Shan, C.; and He, R. 2025.
\newblock {MME}: A Comprehensive Evaluation Benchmark for Multimodal Large
  Language Models.
\newblock In \emph{Advances in Neural Information Processing Systems (NeurIPS),
  Datasets and Benchmarks Track}, volume~38.

\bibitem[{Kahou et~al.(2017)Kahou, Michalski, Atkinson, K{\'a}d{\'a}r,
  Trischler, and Bengio}]{kahou2017figureqa}
Kahou, S.~E.; Michalski, V.; Atkinson, A.; K{\'a}d{\'a}r, {\'A}.; Trischler,
  A.; and Bengio, Y. 2017.
\newblock {FigureQA}: An Annotated Figure Dataset for Visual Reasoning.
\newblock arXiv:1710.07300.

\bibitem[{Kalervo et~al.(2019)Kalervo, Ylioinas, H{\"a}iki{\"o}, Karhu, and
  Kannala}]{kalervo2019cubicasa}
Kalervo, A.; Ylioinas, J.; H{\"a}iki{\"o}, M.; Karhu, A.; and Kannala, J. 2019.
\newblock {CubiCasa5K}: A Dataset and an Improved Multi-task Model for
  Floorplan Image Analysis.
\newblock In \emph{Image Analysis: 21st Scandinavian Conference, SCIA 2019},
  volume 11482 of \emph{Lecture Notes in Computer Science}, 28--40. Springer.

\bibitem[{Kembhavi et~al.(2016)Kembhavi, Salvato, Kolve, Seo, Hajishirzi, and
  Farhadi}]{kembhavi2016ai2d}
Kembhavi, A.; Salvato, M.; Kolve, E.; Seo, M.; Hajishirzi, H.; and Farhadi, A.
  2016.
\newblock A Diagram is Worth a Dozen Images.
\newblock In \emph{Computer Vision -- ECCV 2016}, volume 9908 of \emph{Lecture
  Notes in Computer Science}, 235--251. Springer.

\bibitem[{Kiela et~al.(2021)Kiela, Bartolo, Nie, Kaushik, Geiger, Wu, Vidgen,
  Prasad, Singh, Ringshia, Ma, Thrush, Riedel, Waseem, Stenetorp, Jia, Bansal,
  Potts, and Williams}]{kiela2021dynabench}
Kiela, D.; Bartolo, M.; Nie, Y.; Kaushik, D.; Geiger, A.; Wu, Z.; Vidgen, B.;
  Prasad, G.; Singh, A.; Ringshia, P.; Ma, Z.; Thrush, T.; Riedel, S.; Waseem,
  Z.; Stenetorp, P.; Jia, R.; Bansal, M.; Potts, C.; and Williams, A. 2021.
\newblock Dynabench: Rethinking Benchmarking in {NLP}.
\newblock In \emph{Proceedings of the 2021 Conference of the North American
  Chapter of the Association for Computational Linguistics: Human Language
  Technologies}, 4110--4124. Online: Association for Computational Linguistics.

\bibitem[{Kondratenko et~al.(2026)Kondratenko, Birhane, Hsain, and
  Maciocci}]{kondratenko2026aecvbench}
Kondratenko, A.; Birhane, M.; Hsain, H.~E.; and Maciocci, G. 2026.
\newblock {AECV-Bench}: Benchmarking Multimodal Models on Architectural and
  Engineering Drawings Understanding.
\newblock arXiv:2601.04819.

\bibitem[{Kunz et~al.(2025)Kunz, Klostermeier, Thanabalan, Legler, and
  Ruskowski}]{kunz2025techmb}
Kunz, L.; Klostermeier, M.; Thanabalan, K.; Legler, T.; and Ruskowski, M. 2025.
\newblock {TechMB}: Exploring the Potential of Vision Language Models for
  Interpreting Technical Drawings.
\newblock In \emph{DS 140: Proceedings of the 36th Symposium Design for X
  (DFX2025)}, 179--188. The Design Society.

\bibitem[{Li et~al.(2024{\natexlab{a}})Li, Ge, Ge, Wang, Wang, Zhang, and
  Shan}]{li2023seedbench}
Li, B.; Ge, Y.; Ge, Y.; Wang, G.; Wang, R.; Zhang, R.; and Shan, Y.
  2024{\natexlab{a}}.
\newblock {SEED-Bench}: Benchmarking Multimodal Large Language Models.
\newblock In \emph{Proceedings of the IEEE/CVF Conference on Computer Vision
  and Pattern Recognition (CVPR)}, 13299--13308.

\bibitem[{Li et~al.(2024{\natexlab{b}})Li, Yang, Choi, Zhu, Hsieh, Kim, Lim,
  Ji, Lee, Yan, Petzold, Wilson, Lim, and Wang}]{li2024mmsci}
Li, Z.; Yang, X.; Choi, K.; Zhu, W.; Hsieh, R.; Kim, H.; Lim, J.~H.; Ji, S.;
  Lee, B.; Yan, X.; Petzold, L.~R.; Wilson, S.~D.; Lim, W.; and Wang, W.~Y.
  2024{\natexlab{b}}.
\newblock {MMSci}: A Dataset for Graduate-Level Multi-Discipline Multimodal
  Scientific Understanding.
\newblock arXiv:2407.04903.

\bibitem[{Liang et~al.(2026)Liang, Huang, Wang, Chai, Yu, Wei, Liu, Li, Wang,
  Luo, and Zhao}]{liang2026aecbench}
Liang, C.; Huang, Z.; Wang, H.; Chai, F.; Yu, C.; Wei, H.; Liu, Z.; Li, Y.;
  Wang, H.; Luo, R.; and Zhao, X. 2026.
\newblock {AECBench}: A Hierarchical Benchmark for Knowledge Evaluation of
  Large Language Models in the {AEC} Field.
\newblock \emph{Advanced Engineering Informatics}, 71: 104314.

\bibitem[{Liu et~al.(2024)Liu, Duan, Zhang, Li, Zhang, Zhao, Yuan, Wang, He,
  Liu, Chen, and Lin}]{liu2024mmbench}
Liu, Y.; Duan, H.; Zhang, Y.; Li, B.; Zhang, S.; Zhao, W.; Yuan, Y.; Wang, J.;
  He, C.; Liu, Z.; Chen, K.; and Lin, D. 2024.
\newblock {MMBench}: Is Your Multi-modal Model an All-Around Player?
\newblock In \emph{European Conference on Computer Vision (ECCV)}, 216--233.
  Springer.

\bibitem[{Lu et~al.(2024)Lu, Bansal, Xia, Liu, Li, Hajishirzi, Cheng, Chang,
  Galley, and Gao}]{lu2024mathvista}
Lu, P.; Bansal, H.; Xia, T.; Liu, J.; Li, C.; Hajishirzi, H.; Cheng, H.; Chang,
  K.-W.; Galley, M.; and Gao, J. 2024.
\newblock {MathVista}: Evaluating Mathematical Reasoning of Foundation Models
  in Visual Contexts.
\newblock In \emph{International Conference on Learning Representations
  (ICLR)}.

\bibitem[{Lu et~al.(2022)Lu, Mishra, Xia, Qiu, Chang, Zhu, Tafjord, Clark, and
  Kalyan}]{lu2022scienceqa}
Lu, P.; Mishra, S.; Xia, T.; Qiu, L.; Chang, K.-W.; Zhu, S.-C.; Tafjord, O.;
  Clark, P.; and Kalyan, A. 2022.
\newblock Learn to Explain: Multimodal Reasoning via Thought Chains for Science
  Question Answering.
\newblock In \emph{Advances in Neural Information Processing Systems
  (NeurIPS)}, volume~35.

\bibitem[{Luo et~al.(2025)Luo, Liu, Cheng, Wang, Wang, Wei, Wang, Li, Chai,
  Cheng, Ye, Wang, Zhang, Qiao, Zhang, and Zhao}]{luo2025archcad}
Luo, R.; Liu, Z.; Cheng, T.; Wang, J.; Wang, T.; Wei, X.; Wang, H.; Li, Y.;
  Chai, F.; Cheng, F.; Ye, S.; Wang, W.; Zhang, Y.; Qiao, Y.; Zhang, H.; and
  Zhao, X. 2025.
\newblock {ArchCAD-400K}: A Large-Scale {CAD} Drawings Dataset and New Baseline
  for Panoptic Symbol Spotting.
\newblock In \emph{Advances in Neural Information Processing Systems (NeurIPS),
  Datasets and Benchmarks Track}, volume~38.

\bibitem[{Mankodiya et~al.(2026)Mankodiya, Gallik, Galanos, and
  Mulyar}]{mankodiya2026aecbench}
Mankodiya, H.; Gallik, C.; Galanos, T.; and Mulyar, A. 2026.
\newblock {AEC-Bench}: A Multimodal Benchmark for Agentic Systems in
  Architecture, Engineering, and Construction.
\newblock arXiv:2603.29199.

\bibitem[{Masry et~al.(2022)Masry, Long, Tan, Joty, and
  Hoque}]{masry2022chartqa}
Masry, A.; Long, D.~X.; Tan, J.~Q.; Joty, S.; and Hoque, E. 2022.
\newblock {ChartQA}: A Benchmark for Question Answering about Charts with
  Visual and Logical Reasoning.
\newblock In \emph{Findings of the Association for Computational Linguistics:
  ACL 2022}, 2263--2279. Dublin, Ireland: Association for Computational
  Linguistics.

\bibitem[{Mathew et~al.(2022)Mathew, Bagal, Tito, Karatzas, Valveny, and
  Jawahar}]{mathew2022infographicvqa}
Mathew, M.; Bagal, V.; Tito, R.; Karatzas, D.; Valveny, E.; and Jawahar, C.~V.
  2022.
\newblock {InfographicVQA}.
\newblock In \emph{Proceedings of the IEEE/CVF Winter Conference on
  Applications of Computer Vision (WACV)}, 1697--1706.

\bibitem[{Mathew, Karatzas, and Jawahar(2021)}]{mathew2021docvqa}
Mathew, M.; Karatzas, D.; and Jawahar, C.~V. 2021.
\newblock {DocVQA}: A Dataset for {VQA} on Document Images.
\newblock In \emph{Proceedings of the IEEE/CVF Winter Conference on
  Applications of Computer Vision (WACV)}, 2200--2209.

\bibitem[{Methani et~al.(2020)Methani, Ganguly, Khapra, and
  Kumar}]{methani2020plotqa}
Methani, N.; Ganguly, P.; Khapra, M.~M.; and Kumar, P. 2020.
\newblock {PlotQA}: Reasoning over Scientific Plots.
\newblock In \emph{Proceedings of the IEEE Winter Conference on Applications of
  Computer Vision (WACV)}, 1516--1525.

\bibitem[{Nie et~al.(2020)Nie, Williams, Dinan, Bansal, Weston, and
  Kiela}]{nie2020anli}
Nie, Y.; Williams, A.; Dinan, E.; Bansal, M.; Weston, J.; and Kiela, D. 2020.
\newblock Adversarial {NLI}: A New Benchmark for Natural Language
  Understanding.
\newblock In \emph{Proceedings of the 58th Annual Meeting of the Association
  for Computational Linguistics}, 4885--4901. Online: Association for
  Computational Linguistics.

\bibitem[{Pramanick, Chellappa, and Venugopalan(2024)}]{pramanick2024spiqa}
Pramanick, S.; Chellappa, R.; and Venugopalan, S. 2024.
\newblock {SPIQA}: A Dataset for Multimodal Question Answering on Scientific
  Papers.
\newblock In \emph{Advances in Neural Information Processing Systems (NeurIPS),
  Datasets and Benchmarks Track}, volume~37.

\bibitem[{Roberts et~al.(2024)Roberts, Han, Houlsby, and
  Albanie}]{roberts2024scifibench}
Roberts, J.; Han, K.; Houlsby, N.; and Albanie, S. 2024.
\newblock {SciFIBench}: Benchmarking Large Multimodal Models for Scientific
  Figure Interpretation.
\newblock In \emph{Advances in Neural Information Processing Systems (NeurIPS),
  Datasets and Benchmarks Track}, volume~37.

\bibitem[{Wang et~al.(2024)Wang, Xia, He, Chen, Liu, Zhu, Liang, Wu, Liu,
  Malladi, Chevalier, Arora, and Chen}]{wang2024charxiv}
Wang, Z.; Xia, M.; He, L.; Chen, H.; Liu, Y.; Zhu, R.; Liang, K.; Wu, X.; Liu,
  H.; Malladi, S.; Chevalier, A.; Arora, S.; and Chen, D. 2024.
\newblock {CharXiv}: Charting Gaps in Realistic Chart Understanding in
  Multimodal {LLMs}.
\newblock In \emph{Advances in Neural Information Processing Systems (NeurIPS),
  Datasets and Benchmarks Track}, volume~37.

\bibitem[{Yue et~al.(2024)Yue, Ni, Zhang, Zheng, Liu, Zhang, Stevens, Jiang,
  Ren, Sun, Wei, Yu, Yuan, Sun, Yin, Zheng, Yang, Liu, Huang, Sun, Su, and
  Chen}]{yue2024mmmu}
Yue, X.; Ni, Y.; Zhang, K.; Zheng, T.; Liu, R.; Zhang, G.; Stevens, S.; Jiang,
  D.; Ren, W.; Sun, Y.; Wei, C.; Yu, B.; Yuan, R.; Sun, R.; Yin, M.; Zheng, B.;
  Yang, Z.; Liu, Y.; Huang, W.; Sun, H.; Su, Y.; and Chen, W. 2024.
\newblock {MMMU}: A Massive Multi-discipline Multimodal Understanding and
  Reasoning Benchmark for Expert {AGI}.
\newblock In \emph{Proceedings of the IEEE/CVF Conference on Computer Vision
  and Pattern Recognition (CVPR)}, 9556--9567.

\bibitem[{Yue et~al.(2025)Yue, Zheng, Ni, Wang, Zhang, Tong, Sun, Yu, Zhang,
  Sun, Su, Chen, and Neubig}]{yue2024mmmupro}
Yue, X.; Zheng, T.; Ni, Y.; Wang, Y.; Zhang, K.; Tong, S.; Sun, Y.; Yu, B.;
  Zhang, G.; Sun, H.; Su, Y.; Chen, W.; and Neubig, G. 2025.
\newblock {MMMU-Pro}: A More Robust Multi-discipline Multimodal Understanding
  Benchmark.
\newblock In \emph{Proceedings of the 63rd Annual Meeting of the Association
  for Computational Linguistics (Volume 1: Long Papers)}, 15134--15186. Vienna,
  Austria: Association for Computational Linguistics.

\bibitem[{Zellers et~al.(2018)Zellers, Bisk, Schwartz, and
  Choi}]{zellers2018swag}
Zellers, R.; Bisk, Y.; Schwartz, R.; and Choi, Y. 2018.
\newblock {SWAG}: A Large-Scale Adversarial Dataset for Grounded Commonsense
  Inference.
\newblock In \emph{Proceedings of the 2018 Conference on Empirical Methods in
  Natural Language Processing (EMNLP)}, 93--104. Brussels, Belgium: Association
  for Computational Linguistics.

\bibitem[{Zellers et~al.(2019)Zellers, Holtzman, Bisk, Farhadi, and
  Choi}]{zellers2019hellaswag}
Zellers, R.; Holtzman, A.; Bisk, Y.; Farhadi, A.; and Choi, Y. 2019.
\newblock {HellaSwag}: Can a Machine Really Finish Your Sentence?
\newblock In \emph{Proceedings of the 57th Annual Meeting of the Association
  for Computational Linguistics}, 4791--4800. Florence, Italy: Association for
  Computational Linguistics.

\end{thebibliography}
\end{document}